\documentclass[10pt,twocolumn,letterpaper]{article}

\usepackage{cvpr}              

\usepackage{graphicx}
\usepackage{amsmath}
\usepackage{amssymb}
\usepackage{booktabs}
\usepackage{algorithm}
\usepackage{algorithmic}

\usepackage[pagebackref,breaklinks,colorlinks]{hyperref}

\usepackage[capitalize]{cleveref}
\crefname{section}{Sec.}{Secs.}
\Crefname{section}{Section}{Sections}
\Crefname{table}{Table}{Tables}
\crefname{table}{Tab.}{Tabs.}

\def\cvprPaperID{*****} 
\def\confName{CVPR}
\def\confYear{2023}

\begin{document}

\title{Real-time Multi-Class Helmet Violation Detection Using Few-Shot Data Sampling Technique and YOLOv8}

\author{\thanks{Corresponding author} Armstrong Aboah \& Bin Wang\\
Department of Radiology\\
Northwestern University\\
{\tt\small {armstrong.aboah, bin.wang}@northwestern.edu}
\and
Ulas Bagci \\
Department of Radiology\\
Northwestern University\\
{\tt\small {ulas.bagci}@northwestern.edu}
\and
Yaw Adu-Gyamfi\\
Department of Civil Engineering\\
University of Missouri-Columbia\\
{\tt\small {adugyamfiy}@missouri.edu}
}
\maketitle
\begin{abstract}
Traffic safety is a major global concern. Helmet usage is a key factor in preventing head injuries and fatalities caused by motorcycle accidents. However, helmet usage violations continue to be a significant problem. To identify such violations, automatic helmet detection systems have been proposed and implemented using computer vision techniques. Real-time implementation of such systems is crucial for traffic surveillance and enforcement, however, most of these systems are not real-time. This study proposes a robust real-time helmet violation detection system. The proposed system utilizes a unique data processing strategy, referred to as \textbf{few-shot data sampling}, to develop a robust model with fewer annotations, and a single-stage object detection model, YOLOv8 (You Only Look Once Version 8), for detecting helmet violations in real-time from video frames. Our proposed method won \textbf{7th place} in the 2023 AI
City Challenge, Track 5, with an mAP score of \textbf{0.5861} on experimental validation data. The experimental results demonstrate the effectiveness, efficiency, and robustness of the proposed system. The code for the few-shot data sampling technique is available at \url{https://github.com/aboah1994/few-shot-Video-Data-Sampling.git}.
\end{abstract}

\section{Introduction}
\label{sec:intro}
\begin{figure}
    \centering
    \includegraphics[width=8cm]{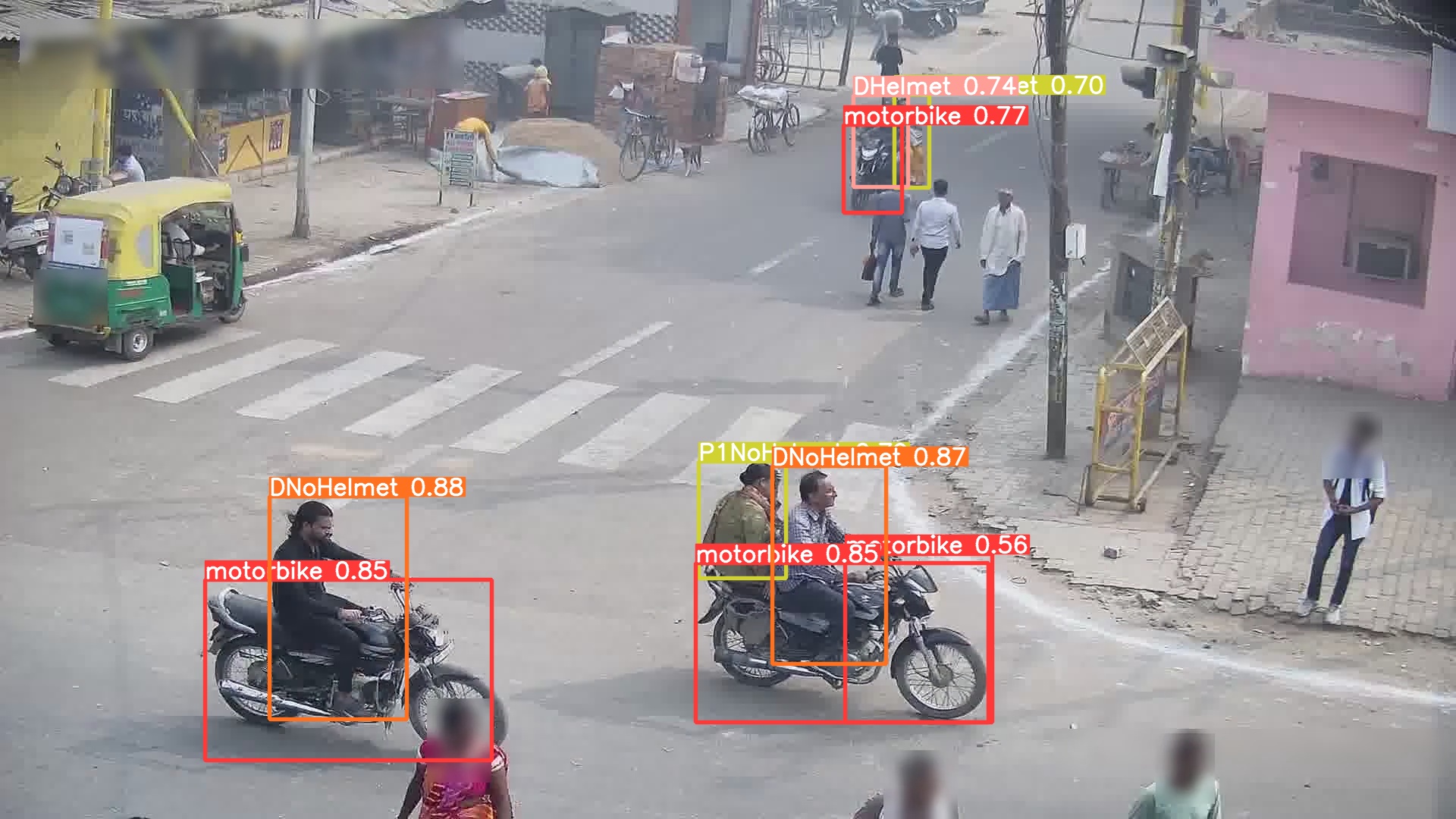}
    \caption{Illustration of predictions of motorists, passengers, and motorcycles in real-time. We present a method that is able to predict helmet rule violation in real-time. The different bounding box colors represent different predictive class.}
    \label{fig:detect}
\end{figure}

Traffic safety is a major concern worldwide, with helmet usage being a key factor in preventing head injuries and fatalities caused by motorcycle accidents. However, helmet usage violations continue to be a significant problem in many countries. In order to detect such violation, various automatic helmet detection systems have been proposed and implemented. These systems use computer vision and machine learning techniques such as object detection, tracking, and recognition to detect and enforce helmet usage violations\cite{zhou2021safety,jia2021real}. Even though several techniques for helmet detection have been proposed in the literature, most of them are not able to perform in real-time. Real-time helmet detection is crucial for traffic surveillance and enforcement, as it allows authorities to quickly identify and take action against individuals who are not wearing helmets (see Fig.~\ref{fig:detect}). This is especially important in developing countries, where helmet usage is not frequent, and the number of motorcycle-related accidents is high.

The need for real-time helmet detection under varying weather conditions and time of day is crucial for ensuring traffic safety. For instance, different weather conditions and lighting can greatly affect the visibility of motorcycle riders and their helmets, making it challenging for traditional systems to accurately detect helmet usage violations. Additionally, the number of helmet usage violations may also vary depending on the time of day. For another example, violations may be more likely to occur during nighttime or in inclement weather when visibility is poor. 

Traditional helmet detection systems often rely on manual inspection which is time-consuming and prone to human error. Real-time helmet detection systems, on the other hand, can automatically and continuously monitor helmet usage in real-time, providing a more efficient and accurate method of enforcing helmet laws. As a result, the current study seeks to develop a real-time helmet violation detection system as shown in Fig.~\ref{fig:detect}.

The overarching goal of this study is to develop a real-time helmet violation detection system that is robust to varying weather conditions and time of day. To achieve this goal, we proposed a unique data processing strategy referred to as the \textbf{\textit{"few-shot data sampling technique"}}. This technique was implemented to develop a more robust model with fewer annotations(see Algorithm~\ref{alg:cat} and \ref{alg:frame}). In our proposed system, we have addressed the challenges of varying lighting and weather conditions by using a newly developed algorithm to classify videos under different conditions, which is significant to address the challenges of sampling evenly from all video types. With the help of our newly developed few-shot data sampling techniques and data augmentation, our proposed system has been able to achieve a high level of accuracy and robustness in detecting helmet usage violations under varying weather conditions and time of day. For developing the detection model, we utilized a single-stage object detection model, YOLOv8, for detecting helmets in real-time. YOLOv8 is a state-of-the-art object detection model that has been shown to achieve high accuracy and speed in real-world applications. One of the main advantages of YOLOv8 is its ability to detect objects in real-time, making it well-suited for our task of detecting helmet usage violations in traffic videos. The major contributions of our study are summarized below; 
\begin{enumerate}
 \item Proposed and implemented a novel data processing strategy, called the \textbf{\textit{"few-shot data sampling technique"}}, for developing a robust helmet detection model with fewer annotations. The technique involves selecting a small but representative number of images from a large dataset using our developed algorithms and then applying data augmentation techniques to generate additional images for training. By using this technique, we able to develop a robust helmet detection model with fewer annotations, which is an important contribution as it reduces the time and effort required for annotation

 \item Developed a real-time helmet detection system that is robust to varying weather conditions and time of day by utilizing YOLOv8, a state-of-the-art object detection model, and data augmentation techniques.  YOLOv8 is designed to be fast and accurate, making it ideal for use in real time. In addition, several data augmentation strategies were utilized in this study to overcome the issues of occlusion and viewpoint problems. To further improve prediction accuracy and confidence during inference time, the study employed test time augmentation (TTA) during its inference stage.

 \item We performed a comparative analysis of three single-stage object detection models from the YOLO series, namely YOLOv5, YOLOv7, and YOLOv8. By evaluating their performance in detecting helmet violations, we aimed to identify the most effective model for detecting helmet violations. This analysis is useful in improving the accuracy and effectiveness of our helmet violation detection system. 

\end{enumerate}

Our experimental results demonstrated the effectiveness of our proposed helmet detection system in detecting helmets in real-world scenarios. The high precision rate achieved in our tests indicates that the system is able to accurately detect helmets in a wide range of visual conditions, including varying weather conditions and different times of day. This is crucial for the practical application of the system, as it ensures that the system can be used to detect helmets in a variety of real-world environments. Secondly, our experimental results showcase the robustness of our proposed data processing strategies. By developing a model that can generalize its predictions to account for many of the visual complexities and situational ambiguities that are often encountered in real-world scenarios, we have created a system that is more robust and reliable than existing methods. Finally, the use of a single-stage object detection model and our proposed data processing strategies have resulted in a model that is both fast and efficient, which is crucial for real-time applications. As such, our experimental results demonstrate that our proposed helmet detection system is not only accurate but also well-suited for real-world use.

The remainder of the paper is structured as follows: In section 2, we conduct a literature review on helmet detection techniques. In section 3, we describe the data set and our unique data processing strategies that were employed to develop a more robust model with fewer annotations. In section 4, we discuss the helmet detection models employed in this study. In section 5, we present our experimental findings, which demonstrate the efficacy of our proposed method in detecting helmets in real-world scenarios. In section 6, we discuss the implications of our findings and make suggestions for future research in this field.

\section{Related works}
\label{sec:lit}

Helmet violation detection, also known as helmet enforcement, has become an important area of research in recent years due to the importance of helmet usage in reducing injury and fatalities in road accidents. In a study conducted by the World Health Organization (WHO), it was estimated that helmets can reduce the risk of death by almost 40\% and the risk of severe injury by 70\% \cite{WHO2013}.

One of the earliest studies in the field of helmet violation detection was conducted by \cite{Zeng2006}, who proposed a method for detecting helmet violations using computer vision techniques. The authors used color and texture-based features to detect helmets in real-time and reported an accuracy rate of 89.5\% \cite{ashvini2017view,wu2018intelligent,wu2019helmet}.

In a more recent study, \cite{Zhang2017} proposed a deep learning-based approach for helmet enforcement in real-time. The authors used a Convolutional Neural Network (CNN) trained on a large dataset of helmet and non-helmet images and reported a high accuracy rate of 97.5\%. The authors also reported that their method was faster than previous methods, with a processing time of less than 50 milliseconds per frame\cite{wu2019helmet,cheng2021multi,deng2022lightweight,song2022multi}.

Another study by \cite{Huang2018} used a deep neural network, specifically YOLOv3, for helmet enforcement in real-time. The authors used a dataset of helmet and non-helmet images to train the network and reported an accuracy rate of 96.2\%. The authors also reported that their method was faster than previous methods, with a processing time of less than 30 milliseconds per frame.

In a recent study, \cite{Gogoi2021} proposed a real-time helmet enforcement system using a combination of computer vision and deep learning techniques. The authors used a combination of color and texture-based features and a deep neural network to detect helmets in real-time and reported an accuracy rate of 95.6\%. The authors also reported that their method was faster than previous methods, with a processing time of less than 100 milliseconds per frame.

In conclusion, helmet violation detection has been an active area of research in recent years, and various studies have proposed different approaches using computer vision and deep learning techniques. The studies mentioned above demonstrate that deep learning-based approaches have achieved high accuracy rates and processing times, making them suitable for real-time helmet enforcement systems. However, more research is needed to improve the accuracy and processing times of these systems and to make them more robust to different conditions and environments.

\section{Data}
\label{sec:data}
\subsection{Data Overview}
\begin{figure}
    \centering
    \includegraphics[width=8cm]{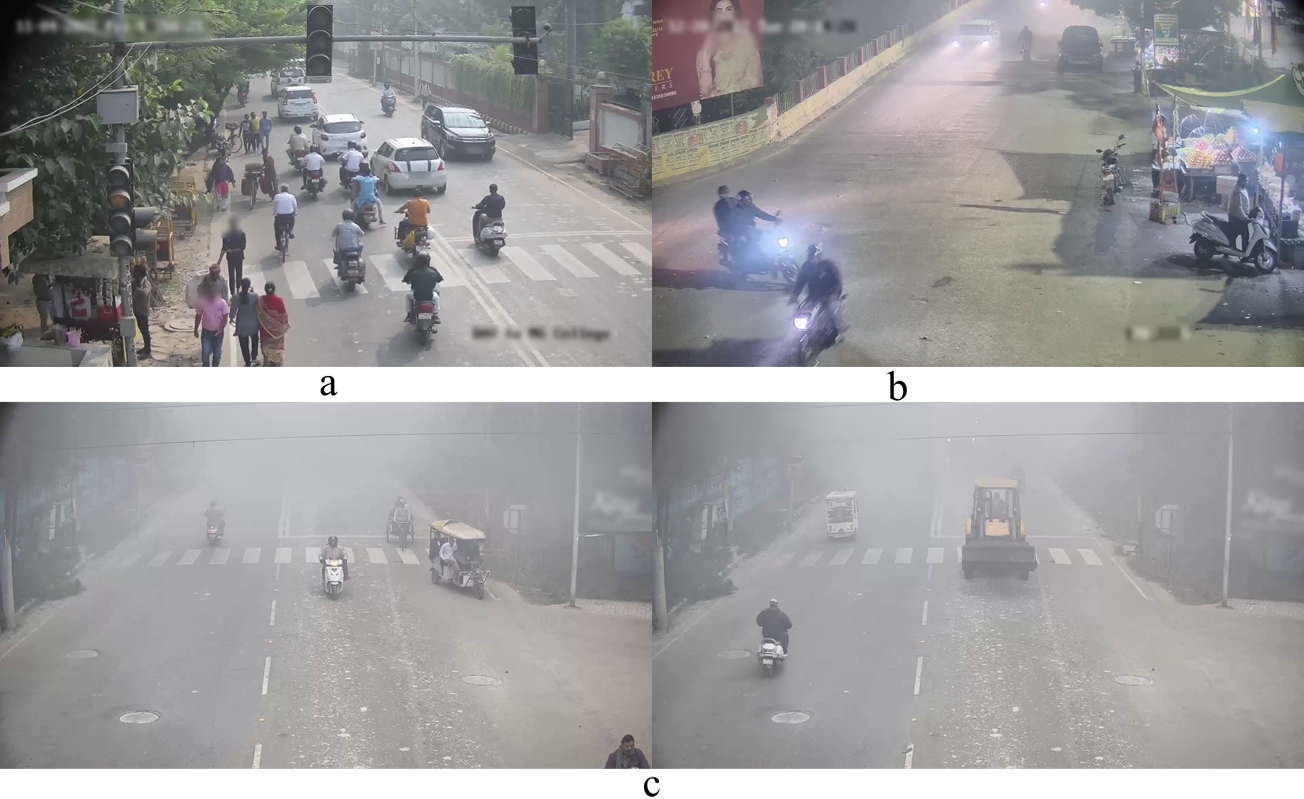}
    \caption{Images from different time of day and weather. a) Daylight, b) Nighttime, and c)Foggy weather.}
    \label{fig:dataset}
\end{figure}

The data used for the proposed violation detection model was sourced from the 2023 NVIDIA AI CITY CHALLENGE, specifically Track 5 which focuses on detecting helmet rule violations among motorcyclists and passengers. The dataset includes 100 videos for training and 100 videos for testing, each with an average length of 20 seconds, a frame rate of 10 fps, and a resolution of 1920x1080. The videos present a range of difficulties, as they feature various camera angles, lighting, and weather conditions as presented in Fig.~\ref{fig:dataset}. The primary objective of the model is to identify motorcycles and their riders and determine whether or not they are wearing helmets.

\subsection{Data Processing}
One of the major steps taken in this study to build a robust helmet violation detection model was to perform certain key data pre-processing steps.

\begin{figure}
    \centering
    \includegraphics[width=8cm]{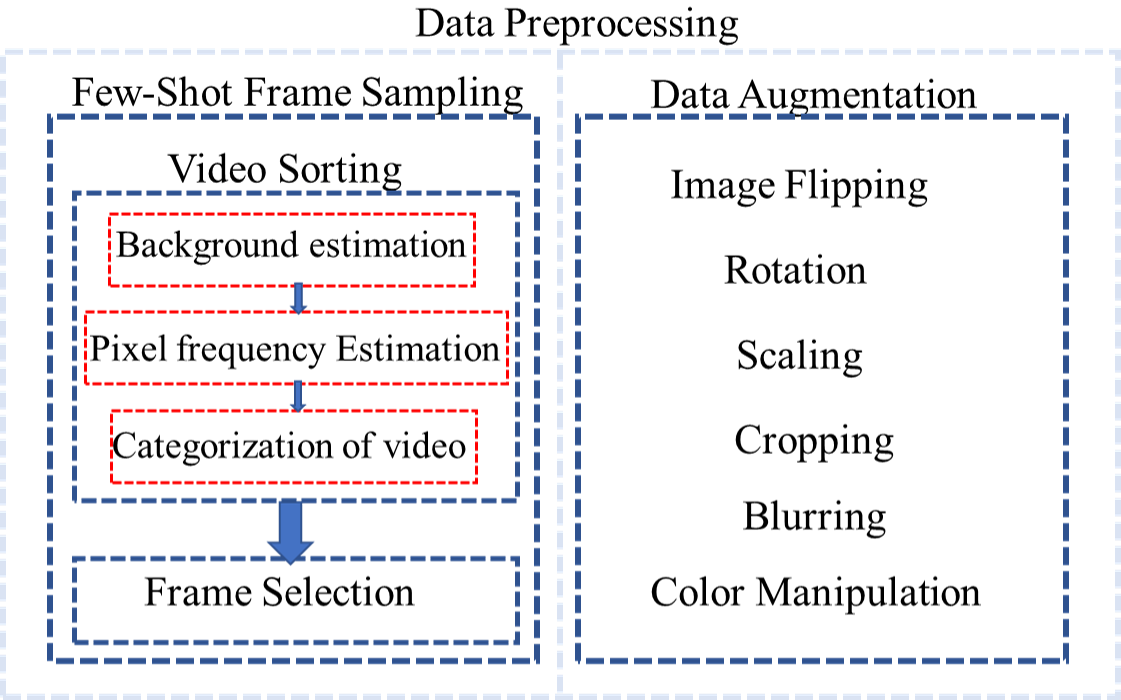}
    \caption{Data Preprocessing: Frame Sampling and Data Augmentation}
    \label{fig:aug}
\end{figure}

The pre-processing steps used in this study are summarized in Fig.~\ref{fig:aug}. The aim of this step was to improve the performance of the object detection model by automatically selecting frames that profoundly represent the training data and also provide enough variability in the training dataset through data augmentation to help the developed model generalize well with real-world test cases. The study carried out two main data pre-processing steps: (1) a few-shot data sampling framework to select the best representative set of data for training and (2) data augmentation to increase the variety of the training data. An illustration of the data pre-processing method, which incorporates six data augmentation techniques, is shown in Fig.~\ref{fig:aug}.

\textbf {Few-shot Data Sampling Framework:} The initial ground truth annotations were provided as part of the challenge, however, there were some missing annotations as illustrated in Fig.~\ref{fig:missed_anno}. These missing annotations have a significant impact when training a model. To address this without manually reviewing all 20,000 frames and correcting annotations, a few-shot data sampling framework was developed. This framework was designed to help select the most representative frames of the entire dataset and minimize the need for re-annotation of all 20,000 frames. The framework consists of three primary steps as follows: 

\begin{figure*}[ht!]
    \centering
    \includegraphics[width=17cm]{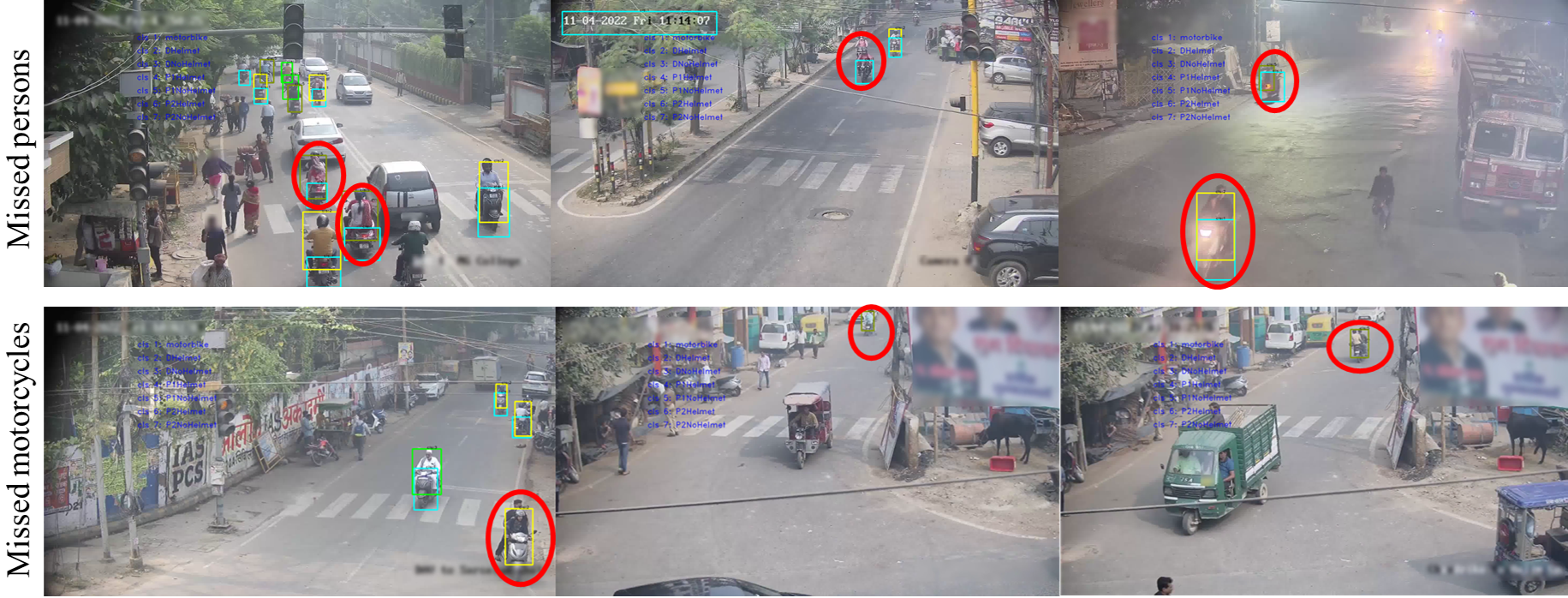}
    \caption{Examples of missed annotations from the ground truth annotations.}
    \label{fig:missed_anno}
\end{figure*}

\textbf{First}, we determine the background in each video. To estimate the background of a video, we first randomly select frames within a 10-second period. Next, we compute the median of 60 percent of all frames in the sample. By using random sampling and determining the median of a subset of images, we are able to negate the impact of short-term video resolution changes such as zooms, and pixelation.

\begin{figure}
    \centering
    \includegraphics[width=8cm]{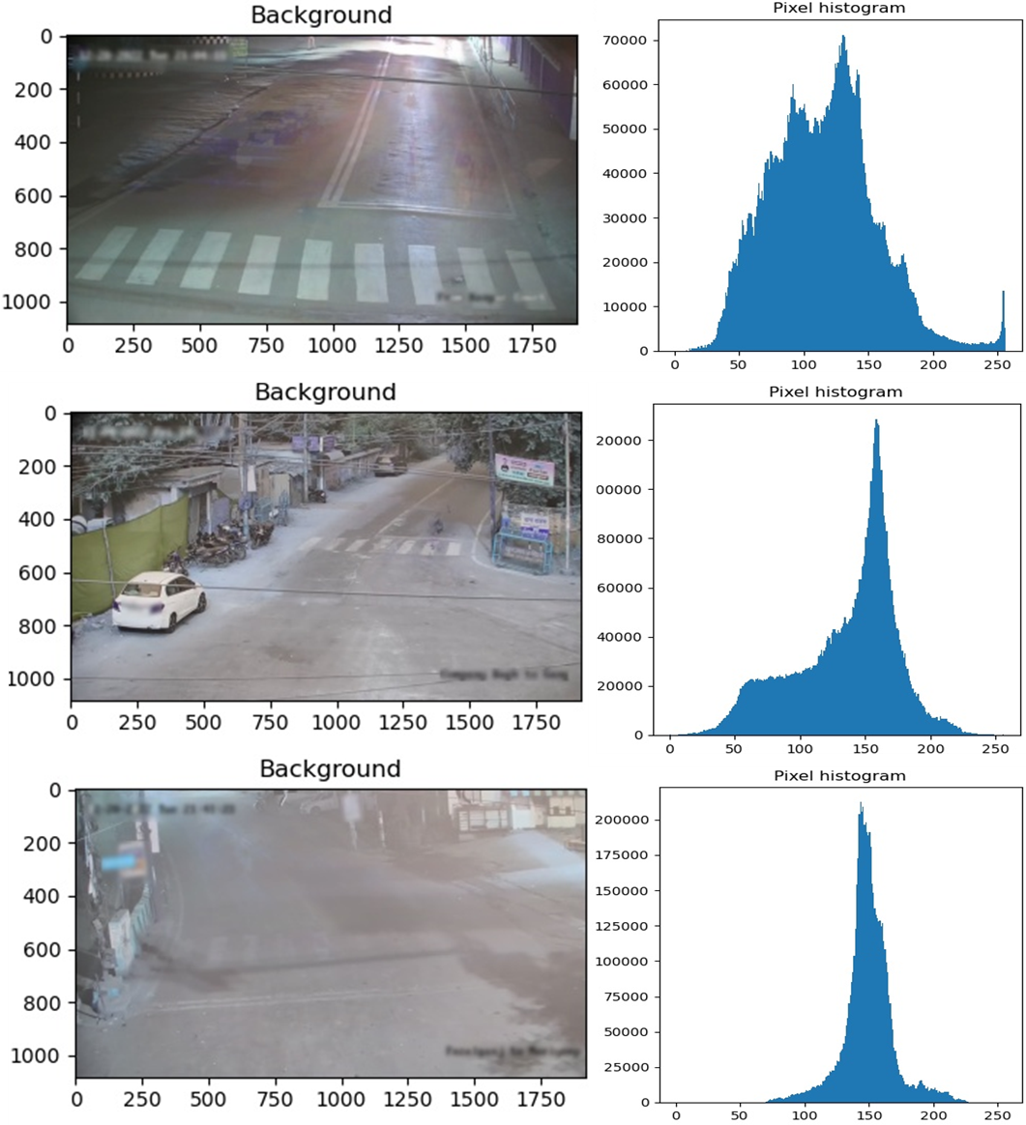}
    \caption{Sorting videos by time of the day and weather condition using frequency thresholding.}
    \label{fig:time}
\end{figure}

\textbf{Second}, our developed algorithm (see Algorithm~\ref{alg:cat}) is used to categorize the videos according to the time of day and weather conditions such as day, night, and fog. It is important to ensure a balanced representation of all video types since there are more daylight videos in the dataset compared to other scenes causing imbalance problems. The proposed algorithm takes the estimated video background and calculates the frequency of each pixel. If the maximum frequency corresponds to a pixel value less than 150, the algorithm classifies the image as night; otherwise, the algorithm classifies the image as day or foggy. To distinguish between daytime and foggy videos, the skewness of the image frequencies is computed. The algorithm classifies the video as foggy if the absolute skewness is close to zero. The frequency distribution of the day, night, and fog images are shown in Fig.~\ref{fig:time}.

\begin{algorithm}
\caption{Categorize Videos}
\label{alg:cat}
\textbf{Input:} Video background image, $I_{bg}$ \\
\textbf{Output:} Categorized video: "day", "night", or "foggy"
\begin{algorithmic}[1]
\FOR{$p \in pixels$}
    \STATE $f_p \gets calculateFrequency(p)$
\ENDFOR
\STATE $maxFrequency \gets max(f_p)$
\IF{$maxFrequency < 150$}
    \STATE $class \gets "night"$
\ELSE
    \STATE $skewness \gets calculateSkewness(f_p)$
    \IF{$|skewness| \approx 0$}
        \STATE $class \gets "foggy"$
    \ELSE
        \STATE $class \gets "day"$
    \ENDIF
\ENDIF
\RETURN $class$
\end{algorithmic}
\end{algorithm}

\textbf{Lastly}, we developed a frame sampling algorithm (see Algorithm~\ref{alg:frame}) that aims to select more frames from video types that were underrepresented, as identified. The algorithm receives as input the anticipated number of frames. With the information regarding the total number of videos in each category and the fps of each, a sample rate is calculated for each video category.

\begin{algorithm}
\caption{Frame Sampling Algorithm}
\label{alg:frame}
\textbf{Input:} $n_{frames}$, $n_{videos}^{category}$, $fps^{category}$\
\textbf{Output:} Selected frames from each video category
\begin{algorithmic}[1]
\FOR{$category \in {1,2,3,...,n}$}
\STATE Calculate sample rate for each category: $sample_rate^{category} = \frac{n_{frames}}{\sum_{i=1}^n n_{videos}^{i} \times fps^{i}}$
\STATE Select frames from each video in the category at the calculated sample rate
\ENDFOR
\STATE Return the selected frames
\end{algorithmic}
\end{algorithm}

\textbf {Data Augmentation:} Data augmentation is a common technique used in computer vision to increase the sample size and diversity of the training dataset. This is particularly critical for object detection tasks such as our current work, which can be complicated by problems of occlusion and changes in viewpoint. To solve the challenges of occlusion and viewpoint variation, several data augmentation techniques were applied in this study. 

\begin{enumerate}
 \item \textbf{Image flipping:} Flipping the image horizontally was done to aid the model to learn to detect helmets from both sides of the motorcycle.

 \item \textbf{Rotation:} Rotation was applied to augment the data by changing the viewpoint angle of the helmet.

 \item \textbf{Scaling:} Scaling was used to change the size of the helmet in the image, which can help the model learn to detect helmets of different sizes.

 \item \textbf{Cropping:} Cropping of images was done to simulate the effect of occlusion, so that the model can learn to detect helmets even when they are partially obscured.

 \item \textbf{Blurring:} Blurring of images were carried out to help the model learn to detect helmets under poor lighting conditions.

 \item \textbf{Color manipulation:} We adjusted the brightness, contrast, and saturation of the image to help the model learn to detect helmets in different lighting conditions.
\end{enumerate}

\subsection{Training and Validation Dataset for Developing the Detection Model}
\begin{figure}[h]
    \centering
    \includegraphics[width=8.3cm]{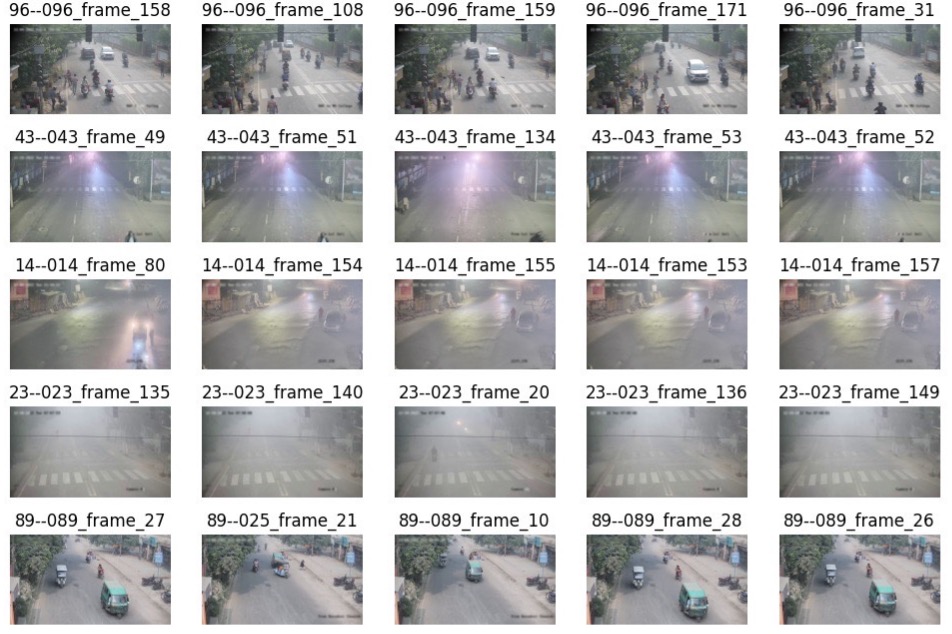}
    \caption{Clustering of frames with high similarity. Each row represent frames of high similarity.}
    \label{fig:sem}
\end{figure}

All models were trained on an NVIDIA GeForce RTX 3090 GPU using 4,500 training examples. The dataset was divided in a ratio of 0.7:0.3 for training and validation respectively. The test dataset was provided separately by the organizers of the competition. To prevent the model from overfitting frames with high similarity, we employed the Semantic Clustering by Adopting Nearest Neighbors (SCAN) algorithm\cite{van2020scan} to eliminate frames with high similarity as shown in Fig~\ref{fig:sem}. In order to eliminate any potential bias in evaluating our validation dataset, we took great care in selecting our training images by ensuring that sequential image sequences did not appear in both the training and validation datasets. Additionally, we added the estimated backgrounds of each video, along with their augmentations, into the training data as negative samples to minimize false positive detections.

\section{Helmet Detection Models}
\label{sec:meth}
The study experimented three object detection models namely YOLOv5, YOLOv7, and YOLOv8. All three models are designed to detect objects in a single stage.

\vspace{0.15in}
\noindent\textbf{YOLOv5}. The YOLOv5\cite{aboah2021vision,aboah2020smartphone,boah2021mobile,Aboah23AIC23,shoman2022region,aboah2023driver} architecture is made up of three main parts which include the Backbone, Neck, and Head. The Backbone has a convolutional neural network that gathers and builds image features at different levels of detail. The Neck combines the features extracted from the Backbone and generates feature maps for the prediction task. The Head uses the feature maps from the Neck to predict object classes and their corresponding bounding boxes. The \textbf{CSPDarknet53} is the backbone network of YOLOv5 which has 29 convolutional layers with a 3x3 kernel size. This results in a receptive field of size 725x725 and a total of 27.6 million parameters. A SPP block is added to the CSPDarknet53 to increase the receptive field without affecting the speed of the operation. Additionally, PANet is used to fuse low-level features with high-level features to enhance the richness of the features.

\vspace{0.15in}
\noindent\textbf{YOLOv7}. The YOLOv7\cite{wang2022yolov7} model had numerous changes to its architecture that distinguished it from YOLOv5. These included the implementation of compound scaling, the extended efficient layer aggregation network (EELAN), and the use of techniques like planned and reparameterized convolution, coarseness for auxiliary loss, and fineness for lead loss. The EELAN was a crucial element of YOLOv7, and it allowed the model to learn more effectively while still maintaining the original gradient route. This was achieved through a technique called "expand, shuffle, and merge cardinality." The compound scaling approach was introduced to YOLOv7 to preserve the model's properties at the initial level and maintain its optimal design. In addition, the paper presented the use of model re-parameterization and dynamic label assignment as network optimization strategies, addressing existing problems and improving the model's overall performance.

\vspace{0.15in}
\noindent\textbf{YOLOv8}. YOLOv8 is the latest iteration in the YOLO family of detection models, which are known for their capabilities for joint detection and segmentation. Similar to YOLOv5\cite{aboah2021vision,shoman2022region}, the architecture consists of a backbone, head, and neck. It boasts a new architecture, improved convolutional layers (backbone), and a more advanced detection head, making it a top choice for real-time object detection. YOLOv8 also offers support for the latest computer vision algorithms such as instance segmentation, enabling multiple object detection in an image or videos. The model uses the \textbf{Darknet-53} backbone network, which is faster and more precise than the previous YOLOv7\cite{wang2022yolov7} network. YOLOv8 predicts bounding boxes through an anchor-free detection head. The model is more effective than previous versions due to its larger feature map and improved convolutional network, which enhances its precision and speed. YOLOv8 also incorporates feature pyramid networks for recognizing objects of varying sizes. The model also comes with a user-friendly API, making it easy to implement in various applications.

\subsection{Model Training}
All three models were trained with optimal hyperparameters generated by genetic algorithm. Table \ref{tab:my-table5} provides a summary of the training hyperparameters. Additionally, all models were trained for 400 epochs with a batch size of 16 and an image size of 832x832.

\begin{table}[h]
\centering
\caption{Training Hyperparaters}
\label{tab:my-table5}
\resizebox{\columnwidth}{!}{%
\begin{tabular}{@{}llllll@{}}
\toprule
\textbf{Hyperparameter}        & \textbf{YOLOv5} & \textbf{YOLOv7} & \textbf{YOLOv8} \\ \midrule
Initial learning rate & 0.0101   & 0.0101    & 0.0106    \\
Optimizer             & Adam     & Adam    & Adam \\
Momentum              & 0.937   & 0.965   & 0.971  \\
Weight decay          & 0.00042  & 0.00051  & 0.00048 \\
Warmup epochs         & 3.862       & 5.541       & 2.689    \\
IoU                   & 0.881     & 0.867     & 0.912 \\
\bottomrule
\end{tabular}}
\end{table}

\subsection{Test Time Augmentation (TTA)}
Test Time Augmentation (TTA) is a technique used in deep learning for improving the accuracy of a model's predictions on test data. TTA involves applying data augmentation techniques, such as rotation, flipping, or cropping, to the test data and then making predictions on each augmented version of the test data. The final prediction is then made by averaging the predictions made on the augmented versions of the test data. TTA can be computationally expensive, as it involves making multiple predictions on augmented versions of the test data. However, it can be implemented efficiently by using parallel processing or by batching the augmented data.

\section{Results and Discussion}
\label{sec:res}

\begin{figure*}[ht!]
    \centering
    \includegraphics[width=17cm]{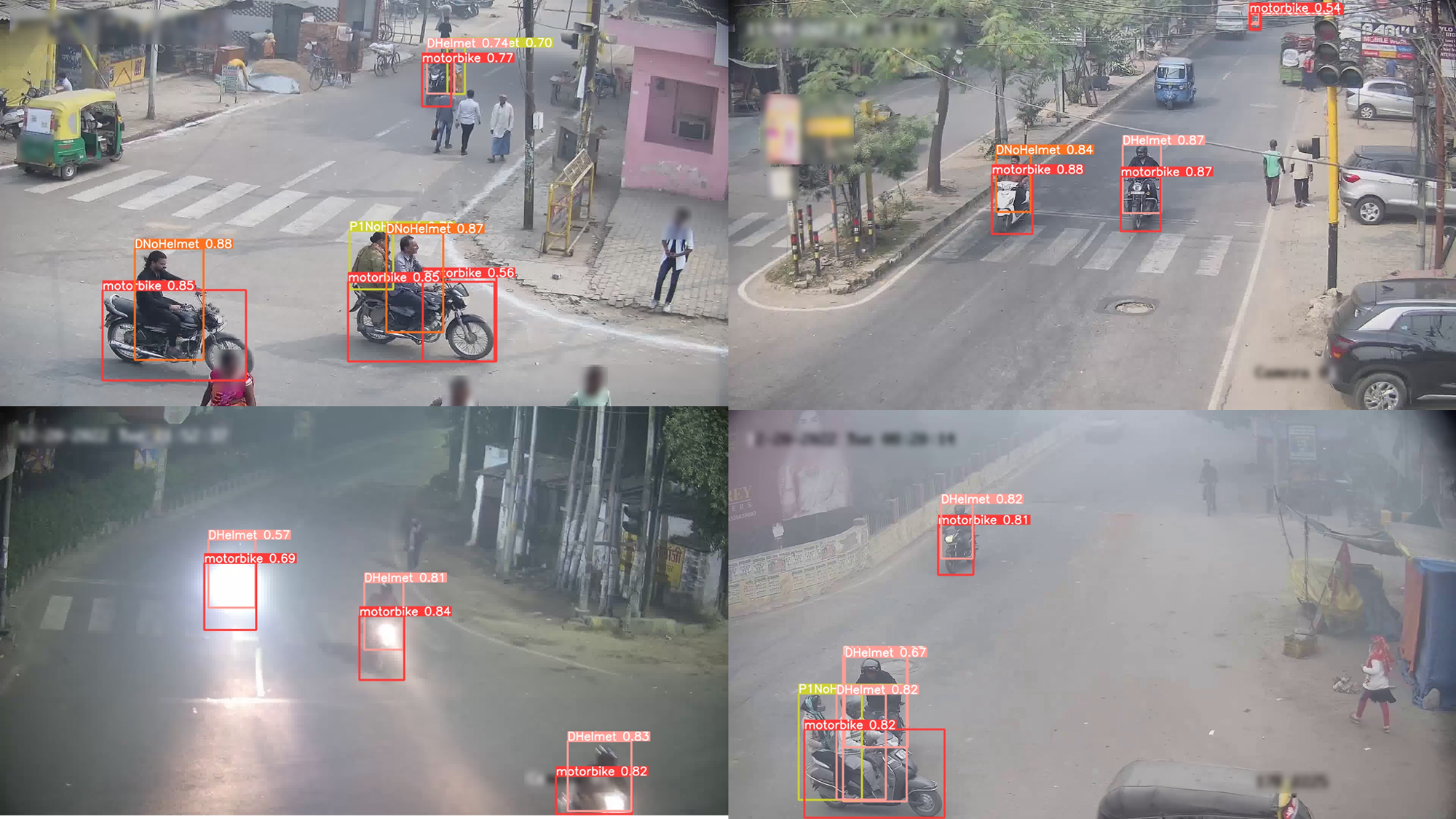}
    \caption{Model detection from different time of day and weather condition.}
    \label{fig:pred_img}
\end{figure*}

The 2023 NVIDIA AI City Challenge Task 5 includes 100 unannotated videos for testing. Each testing video has a 20-second duration and a resolution 1920 × 1080 pixel. Trained models must be able to detect motorcycles and their riders, whether or not they are wearing helmets, from the test videos. Our trained model's detections on the test videos are shown in Fig~\ref{fig:pred_img}.

A submission to the challenge is a text file with the format: Video ID, frame ID, bb\_left, bb\_top, bb\_width, bb\_height, Class ID and Confidence. The Video ID is a numeric identifier for the video, starting from 1 and indicating its position in the alphabetically sorted list of all Task 5, Test Set videos. The frame ID refers to the number of the frame with the detected objects. The bb\_left, bb\_top, bb\_width, and bb\_height are the bounding box coordinates for the detected object and the Class ID is the numeric identifier for the object, starting from 1. The confidence is a probability value between 0 and 1.

The evaluation of our model is based on the mean Average Precision (mAP) across all frames in the test videos, and the leaderboard ranking of submissions is determined by the mAP score, as calculated by Equation ~\ref{eq:map}.
\begin{equation}
\label{eq:map}
    m A P=\frac{1}{N} \sum_{i=1}^N A P_i
\end{equation}

where N is the number of queries.

\subsection{Comparative Analysis}

We conducted a comparative analysis of various single-stage object detection models belonging to the YOLO series with and without test time augmentation (TTA) on both our validation dataset and test dataset(provided by organizers of the competition). 

\subsubsection{Validataion Dataset}
Our analysis presented in Table \ref{tab:my-table} suggests that recent iterations of YOLO, namely yolov7 and yolov8, display superior performance in terms of both mAP.05 and mAP.05-.95 when compared to the older version, yolov5. Moreover, our findings propound that utilizing test time augmentation (TTA) further enhanced the performance of the models. Notably, yolov8+TTA has demonstrated the highest mAP.05-.95 score of 0.647, indicating its ability to detect objects with high accuracy across a range of IoU thresholds. Thus, our results underscore the significance of employing the latest object detection models and incorporating test time augmentation to improve their efficacy.

\begin{table}[]
\caption{Experimental results on validation dataset}
\label{tab:my-table}
\resizebox{\columnwidth}{!}{%
\begin{tabular}{@{}lllll@{}}
\toprule
\textbf{Model}                          & mAP.05                        & mAP.05-.95                    & Precision                     & Recall         \\ \midrule
yolov5 & 0.823 & 0.465 & 0.892 & 0.811          \\
yolov7                         & 0.846                         & 0.526                         & 0.912                         & 0.854          \\
yolov8                                  & 0.858                         & 0.601                         & 0.923                         & 0.898          \\
yolov5+TTA                              & 0.911                          & 0.613                         & 0.931                         & 0.907          \\
\textbf{yolov8+TTA}                     & \textbf{0.923}                & \textbf{0.647}                & \textbf{0.953}                & \textbf{0.918} \\ \bottomrule
\end{tabular}%
}
\end{table}

\subsubsection{Test Dataset}

On the experimental test dataset, the yolov8+TTA outperformed all other models. It achieved an overall mAP score of \textbf{0.5861} with a comparative inference speed of 95 fps as shown in Table \ref{tab:my-table1}. Its inference speed makes it suitable for real-time predictions. The results from on experimental test data was \textbf{ranked 7th} on the public leader board illustrated in Table \ref{tab:my-table2}.

\begin{table}[]
\centering
\caption{Experimental results on test dataset}
\label{tab:my-table1}
\begin{tabular}{@{}lll@{}}
\toprule
Model      & mAP            & fps         \\ \midrule
yolov5     & 0.3988          & 160         \\
yolov7     & 0.4061          & 163         \\
yolov8              & 0.5136          & 162         \\
yolov5+TTA          & 0.5346           & 100         \\
\textbf{yolov8+TTA} & \textbf{0.5861} & \textbf{95} \\ \bottomrule
\end{tabular}
\end{table}

\begin{table}[h]
\centering
\caption{Top 10 Leader Board Ranking}
\label{tab:my-table2}
\begin{tabular}{@{}llll@{}}
\toprule
\textbf{Rank}   & Team ID  & Team Name & Score    \\ \midrule
1 & 58 & CTC\-AI & 0.8340          \\
2 & 33 & SKKU Automation Lab & 0.7754          \\
3 & 37 & SmartVision & 0.6997          \\
4 & 18 & UT\_He & 0.6422   \\
5 & 16 & UT\_NYCU\_SUNY\-Albany & 0.6389         \\
6 & 45 & UT1 & 0.6112         \\
\textcolor{red}{7} & \textcolor{red}{192} & \textcolor{red}{Legends (Ours)} & \textcolor{red}{0.5861}          \\
8 & 55 & NYCU\-ROAD BEAST & 0.5569          \\
9 & 145 & WITAI\-53 & 0.5474          \\
10 & 11 & AIMIZ & 0.5377          \\
 \bottomrule
\end{tabular}%
\end{table}

\section{Conclusion}
\label{sec:con}
We developed a real-time multi-class helmet violation detection system that is robust to varying weather conditions and times of the day. To achieve this goal, we proposed a unique data processing strategy referred to as the \textbf{\textit{"few-shot data sampling technique"}} and utilized YOLOv8, a state-of-the-art single-stage object detection model. Our experimental results demonstrated the effectiveness and robustness of the proposed system in detecting helmets in real-world scenarios with high precision and efficiency. The results also showed that the proposed system is well-suited for practical use in real-time applications.


{\small
\bibliographystyle{unsrt}
\bibliography{egbib}

\begin{thebibliography}{10}

\bibitem{zhou2021safety}
Fangbo Zhou, Huailin Zhao, and Zhen Nie.
\newblock Safety helmet detection based on yolov5.
\newblock In {\em 2021 IEEE International conference on power electronics,
  computer applications (ICPECA)}, pages 6--11. IEEE, 2021.

\bibitem{jia2021real}
Wei Jia, Shiquan Xu, Zhen Liang, Yang Zhao, Hai Min, Shujie Li, and Ye~Yu.
\newblock Real-time automatic helmet detection of motorcyclists in urban
  traffic using improved yolov5 detector.
\newblock {\em IET Image Processing}, 15(14):3623--3637, 2021.

\bibitem{WHO2013}
World Health~Organization (WHO).
\newblock Helmet promotion in low- and middle-income countries, 2013.

\bibitem{Zeng2006}
Z~Zeng, Z~Zhang, and Y~Liu.
\newblock Helmet enforcement using computer vision techniques.
\newblock In {\em Proceedings of the 2006 International Conference on Image
  Processing}, pages 453--456. IEEE, 2006.

\bibitem{ashvini2017view}
M~Ashvini, G~Revathi, B~Yogameena, and S~Saravanaperumaal.
\newblock View invariant motorcycle detection for helmet wear analysis in
  intelligent traffic surveillance.
\newblock In {\em Proceedings of International Conference on Computer Vision
  and Image Processing: CVIP 2016, Volume 2}, pages 175--185. Springer, 2017.

\bibitem{wu2018intelligent}
Hao Wu and Jinsong Zhao.
\newblock An intelligent vision-based approach for helmet identification for
  work safety.
\newblock {\em Computers in Industry}, 100:267--277, 2018.

\bibitem{wu2019helmet}
Fan Wu, Guoqing Jin, Mingyu Gao, HE~Zhiwei, and Yuxiang Yang.
\newblock Helmet detection based on improved yolo v3 deep model.
\newblock In {\em 2019 IEEE 16th International conference on networking,
  sensing and control (ICNSC)}, pages 363--368. IEEE, 2019.

\bibitem{Zhang2017}
Y~Zhang, T~Liu, and L~Shao.
\newblock Deep learning based real-time helmet enforcement system.
\newblock In {\em Proceedings of the 2017 ACM on Multimedia Conference}, pages
  1057--1065. ACM, 2017.

\bibitem{cheng2021multi}
Rao Cheng, Xiaowei He, Zhonglong Zheng, and Zhentao Wang.
\newblock Multi-scale safety helmet detection based on sas-yolov3-tiny.
\newblock {\em Applied Sciences}, 11(8):3652, 2021.

\bibitem{deng2022lightweight}
Lixia Deng, Hongquan Li, Haiying Liu, and Jason Gu.
\newblock A lightweight yolov3 algorithm used for safety helmet detection.
\newblock {\em Scientific reports}, 12(1):10981, 2022.

\bibitem{song2022multi}
Hongru Song.
\newblock Multi-scale safety helmet detection based on rsse-yolov3.
\newblock {\em Sensors}, 22(16):6061, 2022.

\bibitem{Huang2018}
J~Huang, X~Liu, and Ming-Hsuan Yang.
\newblock Real-time helmet enforcement using deep learning.
\newblock In {\em Proceedings of the 2018 ACM on Multimedia Conference}, pages
  1793--1801. ACM, 2018.

\bibitem{Gogoi2021}
Rupjyoti Gogoi and Arindam Bora.
\newblock Real-time helmet enforcement system using computer vision and deep
  learning.
\newblock {\em Journal of Ambient Intelligence and Humanized Computing},
  12(1):65, 2021.

\bibitem{van2020scan}
Wouter Van~Gansbeke, Simon Vandenhende, Stamatios Georgoulis, Marc Proesmans,
  and Luc Van~Gool.
\newblock Scan: Learning to classify images without labels.
\newblock In {\em Computer Vision--ECCV 2020: 16th European Conference,
  Glasgow, UK, August 23--28, 2020, Proceedings, Part X}, pages 268--285.
  Springer, 2020.

\bibitem{aboah2021vision}
Armstrong Aboah.
\newblock A vision-based system for traffic anomaly detection using deep
  learning and decision trees.
\newblock In {\em Proceedings of the IEEE/CVF Conference on Computer Vision and
  Pattern Recognition}, pages 4207--4212, 2021.

\bibitem{aboah2020smartphone}
Armstrong Aboah and Yaw Adu-Gyamfi.
\newblock Smartphone-based pavement roughness estimation using deep learning
  with entity embedding.
\newblock {\em Advances in Data Science and Adaptive Analysis},
  12(03n04):2050007, 2020.

\bibitem{boah2021mobile}
Armstrong Aboah, Michael Boeding, and Yaw Adu-Gyamfi.
\newblock Mobile sensing for multipurpose applications in transportation.
\newblock {\em arXiv preprint arXiv:2106.10733}, 2021.

\bibitem{Aboah23AIC23}
Armstrong Aboah, Bagci Ulas, Rashid~Mussah Abdul, Jakisa~Owor Neema, and Yaw
  Adu-Gyamfi.
\newblock Deepsegmenter: Temporal action localization for detecting anomalies
  in untrimmed naturalistic driving videos.
\newblock In {\em The IEEE Conference on Computer Vision and Pattern
  Recognition (CVPR) Workshops}, June 2023.

\bibitem{shoman2022region}
Maged Shoman, Armstrong Aboah, Alex Morehead, Ye~Duan, Abdulateef Daud, and Yaw
  Adu-Gyamfi.
\newblock A region-based deep learning approach to automated retail checkout.
\newblock In {\em Proceedings of the IEEE/CVF Conference on Computer Vision and
  Pattern Recognition}, pages 3210--3215, 2022.

\bibitem{aboah2023driver}
Armstrong Aboah, Yaw Adu-Gyamfi, Senem~Velipasalar Gursoy, Jennifer Merickel,
  Matt Rizzo, and Anuj Sharma.
\newblock Driver maneuver detection and analysis using time series segmentation
  and classification.
\newblock {\em Journal of Transportation Engineering, Part A: Systems},
  149(3):04022157, 2023.

\bibitem{wang2022yolov7}
Chien-Yao Wang, Alexey Bochkovskiy, and Hong-Yuan~Mark Liao.
\newblock Yolov7: Trainable bag-of-freebies sets new state-of-the-art for
  real-time object detectors.
\newblock {\em arXiv preprint arXiv:2207.02696}, 2022.

\end{thebibliography}
}

\end{document}